\documentclass[letterpaper, 10 pt, conference]{ieeeconf}  

\IEEEoverridecommandlockouts                              

\usepackage{tikz}
\usepackage{siunitx}
\usepackage{arydshln}
\usepackage{subcaption}
\usepackage{amsmath}
\usepackage{amsfonts}
\usepackage{amssymb}
\usepackage{url}
\usepackage{multirow}
\usepackage{multicol}
\usepackage{subcaption}
\usepackage{algorithm}
\usepackage{algpseudocode}
\usepackage[switch]{lineno} 

\newcommand*\patchAmsMathEnvironmentForLineno[1]{
  \expandafter\let\csname old#1\expandafter\endcsname\csname #1\endcsname
  \expandafter\let\csname oldend#1\expandafter\endcsname\csname end#1\endcsname
  \renewenvironment{#1}
     {\linenomath\csname old#1\endcsname}
     {\csname oldend#1\endcsname\endlinenomath}}
\newcommand*\patchBothAmsMathEnvironmentsForLineno[1]{
  \patchAmsMathEnvironmentForLineno{#1}
  \patchAmsMathEnvironmentForLineno{#1*}}
\AtBeginDocument{
\patchBothAmsMathEnvironmentsForLineno{equation}
\patchBothAmsMathEnvironmentsForLineno{align}
\patchBothAmsMathEnvironmentsForLineno{flalign}
\patchBothAmsMathEnvironmentsForLineno{alignat}
\patchBothAmsMathEnvironmentsForLineno{gather}
\patchBothAmsMathEnvironmentsForLineno{multline}
}

\newcommand\submittedtext{%
  \footnotesize This work has been submitted to the IEEE for possible publication. Copyright may be transferred without notice, after which this version may no longer be accessible.}

\newcommand\submittednotice{%
\begin{tikzpicture}[remember picture,overlay]
\node[anchor=south,yshift=10pt] at (current page.south) {\fbox{\parbox{\dimexpr0.65\textwidth-\fboxsep-\fboxrule\relax}{\submittedtext}}};
\end{tikzpicture}%
}

\DeclareMathOperator*{\argmax}{arg\,max}
\DeclareMathOperator*{\argmin}{arg\,min}

\title{\LARGE \bf
  Constraints as Rewards:\\Reinforcement Learning for Robots without Reward Functions
}

\expandafter\ifx\csname ifanonymous\endcsname\relax
\author{Anonymous Authors (Removed for double-anonymous review)}
\else
\author{Yu Ishihara$^{1}$, Noriaki Takasugi$^{1}$, Kotaro Kawakami$^{2}$, Masaya Kinoshita$^{1}$, Kazumi Aoyama$^{1}$
  \thanks{$^{1}$Sony Group Corporation, 1-7-1 Konan Minato-ku, Tokyo, 108-0075, Japan. {\tt\small yu.ishihara@sony.com}}%
  \thanks{$^{2}$Sony Global Manufacturing \& Operations Corporation, 1-7-1 Konan Minato-ku, Tokyo, 108-0075, Japan.}%
}
\fi

\begin{document}

\maketitle
\submittednotice 

\thispagestyle{empty}
\pagestyle{empty}

\expandafter\ifx\csname ifdraft\endcsname\relax
    \documentclass[letterpaper, 10 pt, conference]{ieeeconf}
    \overrideIEEEmargins
    \usepackage{siunitx}
\begin{document}
\fi
\begin{abstract}
	Reinforcement learning has become an essential algorithm for generating complex robotic behaviors.
    However, to learn such behaviors, it is necessary to design a reward function that describes the task, 
    which often consists of multiple objectives that needs to be balanced.
    This tuning process is known as reward engineering and typically involves extensive trial-and-error.
    In this paper, to avoid this trial-and-error process, we propose the concept of Constraints as Rewards (CaR).
    CaR formulates the task objective using multiple constraint functions 
    instead of a reward function and solves a reinforcement learning problem 
    with constraints using the Lagrangian-method. 
    By adopting this approach, different objectives are automatically balanced, 
    because Lagrange multipliers serves as the weights among the objectives. 
    In addition, we will demonstrate that constraints, 
    expressed as inequalities, provide an intuitive interpretation of the optimization target designed for the task.
    We apply the proposed method to the standing-up motion generation task 
    of a six-wheeled-telescopic-legged robot and 
    demonstrate that the proposed method successfully acquires the target behavior, 
    even though it is challenging to learn with manually designed reward functions.
\end{abstract}

\expandafter\ifx\csname ifdraft\endcsname\relax
\end{document}
\fi

\expandafter\ifx\csname ifdraft\endcsname\relax
    \documentclass[letterpaper, 10 pt, conference]{ieeeconf}
    \overrideIEEEmargins
    \usepackage{siunitx}
    \usepackage[dvipdfmx]{graphicx}
\begin{document}
\fi

\section{INTRODUCTION}
Recent advances in the field of reinforcement learning have
enabled robots to generate complex manipulation and locomotion behaviors
that were previously considered challenging~\cite{ICRA2021_Taylor,Nature2023_Kaufmann,CORL2023_Zhuang,ICRA2024_Elliot,ScienceRobotics2019_Jemin}.
However, these successes are not solely due to algorithmic progress;
they also depend on the tuning of reward functions conducted by the authors. 
For example, Taylor et al.~\cite{ICRA2021_Taylor} designed a reward function consisting of 
four different objectives
and tuned the weights among these objectives to achieve locomotion behavior in a bipedal robot
that imitates human movement. 
Currently, there is no established systematic method for this tuning process, 
and therefore success often relies on the individual reward designer's expertise.
For further advancement of robots utilizing reinforcement learning,
we argue that we need to avoid this tuning process, widely known as reward engineering,
and establish a design method that does not heavily rely on individual designers.

In this paper, to avoid the trial-and-error involved in designing a reward function,
we propose a new approach to train robots with reinforcement learning: Constraints as Rewards (CaR).
CaR exclusively uses constraints to solve the reinforcement learning problem.
Specifically, CaR formulates the robot's task solely from constraint functions
and removes the reward function from the reinforcement learning formulation.
By applying this transformation and solving the problem using the Lagrangian-method~\cite{ConvexOpt2004_Boyd},
we can automatically tune the weights among different objectives.
As an algorithm to solve the reinforcement learning problem using CaR,
we propose QRSAC-Lagrangian, an extension of the QRSAC algorithm~\cite{Nature2022_Wurman}.
We show that QRSAC-Lagrangian achieves faster and more stable learning
compared to previous algorithms~\cite{ICRA2024_Elliot,Arxiv2019_Achiam,CoRL2020_Ha}.
Furthermore, to facilitate constraint design, we propose four specific designs of constraint functions.
The proposed constraint functions enable an intuitive interpretation of the task objective.
As a result, we can design each constraint more objectively compared to the design
of a conventional reward function.

\begin{figure}
    \centering
    \includegraphics[width=1.0\columnwidth]{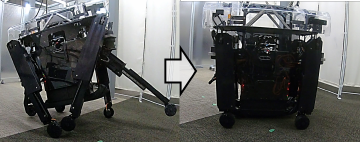}
    \caption{Standing-up motion generation task of a six-wheeled-telescopic-legged 
    \expandafter\ifx\csname ifanonymous\endcsname\relax
    robot.
    \else
    robot: Tachyon 3.
    \fi
    The initial pose (Left) is set randomly, and the robot is requested to transition safely to the upright pose (Right).}
    \label{fig1-1}
\end{figure}

We evaluate the effectiveness of our proposed method by applying it to
the standing-up motion generation task of a six-wheeled-telescopic-legged 
\expandafter\ifx\csname ifanonymous\endcsname\relax
robot 
\else
robot: Tachyon 3~\cite{IROS2024_Takasugi} 
\fi
(Fig.~\ref{fig1-1}).
To enable the robot to stand up, it is necessary to coordinate each leg,
which requires more complex control than conventional quadruped robots.
In addition, the task requires the robot's controller to satisfy multiple conditions
to stand up safely without damaging the hardware.
We show that the task is challenging to learn with manually designed reward functions,
but the proposed method succeeds in acquiring the desired behavior.
Furthermore, we apply the trained policy to the robot in a real environment
and demonstrate that the learned policy successfully achieves the task as it did in the simulation.

In summary, our contributions are as follows:
\begin{itemize}
    \item Proposal of Constraints as Rewards (CaR), a new approach to train robots with reinforcement learning that enables the automatic tuning of weights among objectives by composing the task solely from constraints.
    \item Proposal of four specific designs of constraint function that enable an intuitive interpretation of the task objective.
    \item Proposal of the QRSAC-Lagrangian algorithm, which achieves fast and stable learning in reinforcement learning with CaR.
    \item Evaluation of the effectiveness of the proposed method in both real and simulated environments through the standing-up motion generation task of a six-wheeled-telescopic-legged 
    \expandafter\ifx\csname ifanonymous\endcsname\relax
    robot.
    \else
    robot: Tachyon 3.
    \fi
\end{itemize}

\expandafter\ifx\csname ifdraft\endcsname\relax
\bibliographystyle{IEEEtran}
\bibliography{ref_iros2025}
\end{document}
\fi
\expandafter\ifx\csname ifdraft\endcsname\relax
    \documentclass[letterpaper, 10 pt, conference]{ieeeconf}
    \overrideIEEEmargins
    \usepackage{siunitx}
\begin{document}
\fi

\section{RELATED WORK}
\subsection{Reinforcement Learning with Constraints}
Constraints have been used widely in the field of reinforcement learning to restrict 
the training policy from deviating from the desired state. 
One of the main purposes of constraints is to ensure the safety of the learning policy~\cite{JMLR2015_Garcia}. 
In~\cite{Arxiv2019_Achiam,CoRL2020_Ha}, constraints are employed to define safety requirements, 
and the Lagrangian method is utilized to solve reinforcement learning problems with constraints.
Zhang et al. converted the constraints into penalties and solved an unconstrained optimization problem
to ensure safety~\cite{IJCAI2022_Zhang}.
Liu et al.~\cite{CoRL2022_Liu} constrained the policy to operate in a space tangent to the unsafe region to 
ensure safe exploration.
Constraints are also used for purposes beyond ensuring safety. 
Algorithms such as TRPO~\cite{ICML2015_Schulman} and PPO~\cite{Arxiv2017_Schulman} 
constrain the learning policy from deviation using KL-divergence as a metric. 
In the well-known SAC algorithm~\cite{ICML2018_Haarnoja, Arxiv2019_Haarnoja}, 
constraints were introduced to automatically tune the temperature parameter.
Similar to our work, Elliot et al. proposed Constraints as Terminations (CaT)~\cite{ICRA2024_Elliot}, 
which transform constraints into episode terminations.
However, the problem formulation in previous research still focus on maximizing cumulative rewards 
and require manual tuning of the weights that balance different task objectives. 
In this paper, we propose composing tasks solely from constraints 
to avoid reward engineering and to automatically tune the weights among task objectives.

\subsection{Deep Reinforcement Learning for Legged Robots}
There is no doubt that recent progress in generating complex locomotion behavior in legged robots 
has been achieved through deep reinforcement learning techniques. 
Taylor et al.~\cite{ICRA2021_Taylor} succeessfully retargeted human motion capture data to 
a small bipedal robot. Similarly, Jemin et al.~\cite{ScienceRobotics2019_Jemin} succeeded in 
making a quadruped robot stand up from a fallen pose.
A quadruped robot can now perform parkour using a controller trained with reinforcement learning~\cite{CORL2023_Zhuang}. 
Furthermore, various studies have demonstrated the effectiveness of reinforcement learning~\cite{Arxiv2019_Haarnoja,ICRA2024_Elliot,Arxiv2023_Lee,Science2024_Haarnoja}.
However, we assert that there may have been a significant trial-and-error process involved in reward tuning 
behind these successes, which is often not described in the papers. 
In this work, we propose a new approach to compose the task objective to mitigate this trial-and-error process. 
We applied the proposed approach to the standing-up motion generation task of 
a six-wheeled-telescopic-legged 
\expandafter\ifx\csname ifanonymous\endcsname\relax
robot,
\else
robot, Tachyon 3~\cite{IROS2024_Takasugi},
\fi
and demonstrated its effectiveness.

\expandafter\ifx\csname ifdraft\endcsname\relax
\bibliographystyle{IEEEtran}
\bibliography{ref_iros2025}
\end{document}
\fi
\expandafter\ifx\csname ifdraft\endcsname\relax
    \documentclass[letterpaper, 10 pt, conference]{ieeeconf}
    \overrideIEEEmargins
    \usepackage{siunitx}
    \usepackage{amsmath}
    \usepackage{amsfonts}
    \usepackage{amssymb}
    \DeclareMathOperator*{\argmax}{arg\,max}
    \DeclareMathOperator*{\argmin}{arg\,min}
\begin{document}
\fi

\section{PRELIMINARIES}
In this research, we consider the finite-horizon reinforcement learning problem~\cite{RL2018_Sutton}.
The objective of the problem is to find an optimal policy $\pi^{\ast}$ that maximizes 
the sum of discounted rewards over the horizon $T$:
\begin{equation}
	\label{eq2-1}
	\max_{\pi} \mathbb{E}_{\pi}[\sum_{t=0}^{T}\gamma^{t}r(s_t, a_t)].
\end{equation}
Here, $s_t$, $a_t$, $\gamma$, and $r$ denote 
the state at time $t$, the action at time $t$, the discount factor, and the reward function, respectively.
In this work, we revisit the design of the reward function $r(s,a)$.

As previously mentioned, in many practical applications, 
the reward function $r(s,a)$ is designed as a weighted sum of
multiple functions $r_{n}(s,a)\ (n=1,\dots,N)$
\begin{equation}
	\label{eq2-2}
    r(s,a)=\sum_{n=1}^{N}w_{n}r_{n}(s,a).
\end{equation}
Therefore, the actual problem that needs to be solved is:
\begin{equation}
	\label{eq2-3}
    \max_{\pi}\sum_{n=1}^{N}w_{n}\mathbb{E}_{\pi}\biggl[\sum_{t=0}^{T}\gamma^{t}r_{n}(s_{t}, a_{t})\biggr].
\end{equation}
From this equation, we can confirm that 
we need to tune both weights $w_n$ and functions $r_{n}$ for successful learning. 
However, there is no systematic procedure to tune these parameters. 

In this work, to alleviate the trial-and-error involved in the design of the reward function, 
we consider solving reinforcement learning problem with constraints. 
The reinforcement learning problem with constraints is defined as follows:
\begin{equation}
    \label{eq2-4}
    \begin{split}
    \max_{\pi}\   & \mathbb{E}_{\pi}\biggl[\sum_{t=0}^{T}\gamma^{t}r(s_{t}, a_{t})\biggr]                 \\
    \text{s.t.}\  & \mathbb{E}_{\pi}\biggl[\sum_{t=0}^{T}\gamma^{t}g_{m}(s_t, a_t)\biggr] \geq 0,\ m=1,\dots,M.    
\end{split}
\end{equation}
Here, $g_{m}(s_t, a_t)$ is the $m$-th constraint function.
In the next section, we will show that 
we can eliminate the tuning process required in the original problem using 
this formulation.

\expandafter\ifx\csname ifdraft\endcsname\relax
	\bibliographystyle{IEEEtran}
	\bibliography{ref_iros2025}
\end{document}
\fi
\expandafter\ifx\csname ifdraft\endcsname\relax
    \documentclass[letterpaper, 10 pt, conference]{ieeeconf}
    \overrideIEEEmargins
    \usepackage{siunitx}
    \usepackage[dvipdfmx]{graphicx}
    \usepackage{amsmath}
    \usepackage{amsfonts}
    \usepackage{amssymb}
    \usepackage{algorithm}
    \usepackage{algpseudocode}
\begin{document}
\fi

\section{METHOD}
\subsection{Constraints as Rewards (CaR)}
In this section, we will describe in detail the idea of Constraints as Rewards (CaR).
In CaR, we transform the reinforcement learning problem with constraints described in eq. (\ref{eq2-4})
into an unconstrained problem by incorporating the Lagrange dual function $L(\pi,\boldsymbol{\lambda})$~\cite{ConvexOpt2004_Boyd}:
\begin{equation}
    \label{eq3-1}
    \max_{\lambda_{m}>0\ \forall{m}}\max_{\pi} L(\pi,\boldsymbol{\lambda}).\\
\end{equation}
Where $\boldsymbol{\lambda}=[\lambda_1,\dots,\lambda_{M}]^{\mathsf{T}}$ are the Lagrange multipliers, and
\begin{equation}
    \label{eq3-2}
    \begin{split}
        &L(\pi,\boldsymbol{\lambda})\\
        &\triangleq\mathbb{E}_{\pi}\biggl[\sum_{t=0}^{T}\gamma^{t}r(s_{t}, a_{t})\biggr]
        +\sum_{m=1}^{M}\lambda_{m}\mathbb{E}_{\pi}\biggl[\sum_{t=0}^{T}\gamma^{t}g_{m}(s_t, a_t)\biggr].
    \end{split}
\end{equation}
The transformed problem is known as the Lagrange dual of the original problem~\cite{ConvexOpt2004_Boyd}.
The core idea of CaR is to set the reward function in eq. (\ref{eq3-2}) to $r(s_t,a_t)\triangleq0$.
By doing so, we obtain the following problem to solve:
\begin{equation}
    \label{eq3-3}
    \max_{\lambda_{m}>0\ \forall{m}}\max_{\pi} \sum_{m=1}^{M}\lambda_{m}\mathbb{E}_{\pi}\biggl[\sum_{t=0}^{T}\gamma^{t}g_{m}(s_t, a_t)\biggr].\\
\end{equation}
We can see that this formulation is identical to the standard reinforcement learning problem with a reward function (eq. (\ref{eq2-3}))
except for the operator $\max_{\lambda_{m}>0\ \forall{m}}$.
The new formulation suggests that Lagrange multipliers,
which serves as weights among constraints, can be tuned automatically.
Therefore, if we design the learning objective in terms of constraints,
we can obtain the desired policy without tuning the weights among different objectives.

\subsection{Constraint Function Design}
To compose the learning objective with constraints,
we propose the following four designs of constraint function $g(s,a)$.
Each design provides an intuitive interpretation for the optimization target.
Timestep constraints enable constraining the robot at specific timestep (e.g., constraining the final pose of the robot.),
while episode constraints enable constraining the robot during an episode (e.g., constraining the robot from hitting an obstacle.).
We expect that having an intuitive interpretation will make it easier for the task designer to compose the objective.
In our experiments, we composed the task objective using a combination of these functions.
Please note that the inequality in each constraint can be reversed by setting the constraint function to
$-g(s,a)$.  See the appendix section for the derivation of each function.

\begin{enumerate}
    \item Timestep probability constraint: Constrain the probability of an event at specific timestep $t=t'$ to be less than or equal to $p_{\epsilon}\in[0\dots1]$:
          $p_{\epsilon} \geq P_{\pi}(s_{t'}\in\mathcal{S}',a_{t'}\in\mathcal{A}')$. With:
          \begin{equation}
              \label{eq3-4}
              g(s,a) =
              \begin{cases}
                  0 \quad (t \neq t') \\
                  p_{\epsilon}-\mathbf{1}_{s\in\mathcal{S}',a\in\mathcal{A}'} \quad (t=t').
              \end{cases}
          \end{equation}
    \item Timestep value constraint: Constrain the value computed from a state and/or action in expectation at specific timestep $t=t'$ to be less than or equal to $\epsilon$:
          $\epsilon \geq \mathbb{E}_{\pi}[\hat{g}(s_{t'},a_{t'})]$. With:
          \begin{equation}
              \label{eq3-5}
              g(s,a) =
              \begin{cases}
                  0 \quad (t \neq t') \\
                  \epsilon-\hat{g}(s,a) \quad (t=t').
              \end{cases}
          \end{equation}
    \item Episode probability constraint: Constrain the probability of an event during an episode to be less than or equal to $p_{\epsilon}\in[0\dots1]$:
          $p_{\epsilon} \geq P_{\pi,\gamma}(s\in\mathcal{S}',a\in\mathcal{A}')$. With:
          \begin{equation}
              \label{eq3-6}
              g(s,a) = p_{\epsilon}-\mathbf{1}_{s\in\mathcal{S}',a\in\mathcal{A}'}.
          \end{equation}
    \item Episode value constraint: Constrain the value computed from a state and/or action during an episode in expectation to be less than or equal to $\epsilon$:
          $\epsilon \geq \mathbb{E}_{\pi,\gamma}[\hat{g}(s,a)]$. With:
          \begin{equation}
              \label{eq3-7}
              g(s,a) = \epsilon-\hat{g}(s,a).
          \end{equation}
\end{enumerate}
Here, $\mathcal{S}'$ and $\mathcal{A}'$ are the sets of events of particular interest,
$\mathbf{1}$ is the indicator function, $P_{\pi}$ and $P_{\pi, \gamma}$ are
the undiscounted and discounted state-action probabilities, and
$\mathbb{E}_{\pi,\gamma}$ is the expectation under the discounted state-action distribution.

\subsection{QRSAC-Lagrangian}
We propose QRSAC-Lagrangian\footnote{We designed the algorithm to be applicable to the general problem setting where $r(s_t,a_t)\neq0$.}, an extension of QRSAC~\cite{Nature2022_Wurman},
to solve the reinforcement learning problem when using CaR.
We extended QRSAC because we expect that the quantile function,
which estimates the distribution of Q values,
is effective in our problem setting.
In CaR, the algorithm needs to periodically update the Lagrange multipliers.
Therefore, the distribution of the target Q value changes during training.
In such situations, direct estimation of Q values performed using conventional algorithms may become unstable.
We will compare the performance of QRSAC-Lagrangian, when used in conjunction with CaR,
with variants of Lagrangian-based algorithms, such as SAC-Lagrangian~\cite{CORL2023_Zhuang} and PPO-Lagrangian~\cite{Arxiv2019_Achiam}, and
show that QRSAC-Lagrangian achieves faster convergence compared to these algorithms.
The pseudo code of the algorithm is shown in Algorithm~\ref{alg3-1}. In Algorithm~\ref{alg3-1}, $\pi_{\theta}$ is the training policy,
$d$ is the multiplier update interval, and $p(s_{t+1}| a_t,s_t)$ is the state transition distribution.

\begin{algorithm}
    \caption{QRSAC-Lagrangian}
    \label{alg3-1}
    \begin{algorithmic}[1]
        \State {Initialize policy parameters $\theta$, set replay buffer to $\mathcal{D}=\{\}$, and set Lagrange multipliers $\boldsymbol{\lambda}=\mathbf{0}$}.
        \For {each iteration $i$}
        \State {$a_t \sim \pi_{\theta}(a_t|s_t)$}
        \State {$s_{t+1} \sim p(s_{t+1}| a_t,s_t)$}
        \State {$\mathcal{D}\leftarrow \mathcal{D}\ \cup \ \{(s_t,a_t,r(s_t,a_t),g_{1,\dots,M}(s_t, a_t),s_{t+1})\}$}
        \State {Update policy $\pi_{\theta}$ with QRSAC using $\mathcal{D}$ and $\boldsymbol{\lambda}$}
        \If {$i \bmod d$}
        \For {each $m$}
        \State {$\lambda_{m} \leftarrow \lambda_{m} - Adam(\alpha_{\lambda}, \nabla_{\lambda_{m}}L(\pi, \boldsymbol{\lambda}))$}
        \State {$\lambda_{m} \leftarrow \max(\lambda_{m}, 0)$}
        \EndFor
        \EndIf
        \EndFor
    \end{algorithmic}
\end{algorithm}

\expandafter\ifx\csname ifdraft\endcsname\relax
\bibliographystyle{IEEEtran}
\bibliography{ref_iros2025}
\end{document}
\fi
\expandafter\ifx\csname ifdraft\endcsname\relax
    \documentclass[letterpaper, 10 pt, conference]{ieeeconf}
    \overrideIEEEmargins
    \usepackage{siunitx}
    \usepackage[dvipdfmx]{graphicx}
    \usepackage{url}
    \usepackage{multirow}
    \usepackage{subcaption}
    \usepackage{amssymb}
    \usepackage{amsmath}
    \usepackage{amsfonts}
    \usepackage{siunitx}
\begin{document}
\fi

\section{IMPLEMENTATION}
Our robot has six telescopic legs, four with driving wheels and two with omnidirectional passive wheels.
Each leg has a hip joint with a range of motion of \SI{45}{deg} and a \SI{500}{mm} expandable
prismatic knee joint.
The policy must coordinate each leg to enable the robot to stand up during the task.
In this section, we will describe the design of the constraint functions used in training
and the controller design for this robot.
\expandafter\ifx\csname ifanonymous\endcsname\relax
\else
For hardware details of the robot, please refer to \cite{IROS2024_Takasugi}.
\fi
\subsection{Constraint Function Design for Standing Up Task}
We designed the following five constraints for the task:
\begin{figure}
    \centering
    \includegraphics[width=0.4\columnwidth]{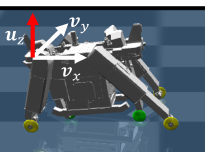}
    \caption{Relationship between $\mathbf{u}_{z}$, $\mathbf{v}_{x}$ and $\mathbf{v}_{y}$.}
    \label{fig4-1}
\end{figure}
\begin{enumerate}
    \item Final pose constraint:
          \begin{equation}
              g(s,a) =
              \begin{cases}
                  0 \quad (t \neq T) \\
                  10^{-3}-|s_{\text{target}} - s_{\text{angle/position}}| \quad (t=T).
              \end{cases}
          \end{equation}
          Existing research~\cite{ScienceRobotics2019_Jemin} uses an episodic reward that
          penalizes the robot's pose at every timestep to facilitate standing up.
          However, from the perspective of constraints, the intermediate pose of the robot should be arbitrary.
          Therefore, we constrain the hip joint angle
          and knee position of the robot's leg only at the final timestep $T$ to be below \SI{e-3}{rad} and \SI{e-3}{m}.
          $s_{\text{target}}$ is the target joint angle/knee position
          and $s_{\text{angle/position}}$ is the actual joint angle/knee position.
          We apply this constraint to each leg, resulting in a total of 12 pose constraints.
    \item Fall down constraint:
          \begin{equation}
              g(s,a) =
              \begin{cases}
                  0 \quad (t \neq T) \\
                  -\mathbf{1}_{\text{robot is in a fall-down state.}} \quad (t=T).
              \end{cases}
          \end{equation}
          This function constrains the probability of falling down to 0.
          The robot is considered to be in a fall-down state whenever more than five wheel joints are higher than the hip joints.
          The task terminates when the robot falls down:
          therefore, the indicator function is evaluated at $t=T$.
    \item Body contact constraint:
          \begin{equation}
              g(s,a) = -\mathbf{1}_{\text{robot's body hits the floor.}}.
          \end{equation}
          This function constrains the probability of the robot's body making contact with the floor to 0.
    \item Leg swing constraint:
          \begin{equation}
              g(s,a) = -\mathbf{1}_{\text{Angular velocity exceeds 2.0 rad/s.}}
          \end{equation}
          This function constrains the angular velocity of the robot's hip joint to be less than \SI{2.0}{rad/s},
          preventing dangerous leg swinging actions.
          To evaluate this constraint, we terminated the episode whenever the joint's angular velocity exceeds \SI{2.0}{rad/s}.
    \item Inclination constraint:
          \begin{equation}
              g(s,a) =
              \begin{cases}
                  0 \quad (t \neq T) \\
                  10^{-2}-|\mathbf{u}_{z}^{\mathsf{T}}\mathbf{v}_{x,y}| \quad (t=T).
              \end{cases}
          \end{equation}
          We add this constraint to ensure that the final pose of the robot remains parallel to the floor.
          We constrain the robot's body to be perpendicular to the z-axis.
          Here, $\mathbf{u}_{z}=[0, 0, 1]^{\mathsf{T}}$ is the unit vector along the z-axis, and
          $\mathbf{v}_{x}$ and $\mathbf{v}_{y}$ are the normalized vectors connecting the leg joints (Fig.~\ref{fig4-1}).
\end{enumerate}

\subsection{Robot Controller Design}
\begin{table}[t]
    \caption{Network structure of $\pi_{\theta}$}
    \label{tab4-1}
    \centering
    \begin{tabular}{|c|c|}
        \hline
        Input layer    & state input                                  \\
        \hline
        Middle layer 1 & Fully-connected followed by ReLU (256 dim)   \\
        \hline
        Middle layer 2 & Fully-connected followed by ReLU (256 dim)   \\
        \hline
        Middle layer 3 & Fully-connected followed by ReLU (256 dim)   \\
        \hline
        Output layer   & Gaussian mean (12 dim) and variance (12 dim) \\
        \hline
    \end{tabular}
\end{table}
Our robot is controlled with a PID controller that receives the target joint angles and positions of each leg.
Therefore, we trained a controller $\pi_{\theta}$ that outputs the target joint angles and positions to be input to this PID-controller.
Table \ref{tab4-1} shows its network architecture.
The policy inputs a vector consisting of the following features:
\begin{itemize}
    \item Hip joint angles (6 dim)
    \item Knee joint positions (6 dim)
    \item Hip joint angular velocities (6 dim)
    \item Knee joint velocities (6 dim)
    \item Robot's angular velocities and accelarations (6 dim)
    \item Timestep since the beginning of the task (1 dim)
    \item History of policy's output (12 dim$\times H$ steps).
\end{itemize}
The output of the policy is squashed and normalized using the tanh function to fit within the range of $[-1,1]$.
Therefore, we rescaled the policy's output before feeding it to the PID-controller.
We fixed the robot's wheel velocity to 0 for the standing-up task. 
We set $H=3$ in the simulation and $H=5$ in the real robot experiment.

\expandafter\ifx\csname ifdraft\endcsname\relax
\bibliographystyle{IEEEtran}
\bibliography{ref_iros2025}
\end{document}
\fi
\expandafter\ifx\csname ifdraft\endcsname\relax
    \documentclass[letterpaper, 10 pt, conference]{ieeeconf}
    \overrideIEEEmargins
    \usepackage{siunitx}
    \usepackage{url}
    \usepackage[dvipdfmx]{graphicx}
    \usepackage{arydshln}
    \usepackage{multirow}
    \usepackage{multicol}
    \usepackage{subcaption}
    \usepackage{amssymb}
    \usepackage{amsmath}
    \usepackage{amsfonts}
\begin{document}
\fi

\section{EXPERIMENTS}
\subsection{Experimental Setup}
\begin{table}[t]
    \caption{Reward functions for comparison. $\boldsymbol{p}_t$ denotes a vector at time $t$
        that consists of joint angles, joint angular velocities, knee positions, and knee velocities of each leg.
        The target value of $\boldsymbol{p}$ at upright pose is $\boldsymbol{p}=\bf{0}$.}
    \label{tab6-1}
    \centering
    \begin{tabular}{l|l}
        \hline
        \multicolumn{2}{c}{Reward design 1}                                \\
        \hline
        Pose reward       & $
            \begin{cases}
                1/(||\boldsymbol{p}_{t}||+1)\quad(t=T) \\
                0\quad(t\neq T)
            \end{cases}
        $                                                                  \\
        Fall down penalty & -1                                             \\
        \hline
        \multicolumn{2}{c}{Reward design 2}                                \\
        \hline
        Pose penalty      & $-w\times||\boldsymbol{p}_{t}||\quad(w=0.001)$ \\
        Fall down penalty & -1                                             \\
        \hline
        \multicolumn{2}{c}{Reward design 3}                                \\
        \hline
        Pose reward       & $
            \begin{cases}
                1/(||\boldsymbol{p}_{t}||+1)\quad(t=T) \\
                1/(||\boldsymbol{p}_{t}||+1)-1/(||\boldsymbol{p}_{t-1}||+1)\quad(t\neq T)
            \end{cases}
        $                                                                  \\
        Fall down penalty & -1                                             \\
        \hline
        \multicolumn{2}{c}{Reward design 4}                                \\
        \hline
        Pose reward       & $1/(||\boldsymbol{p}_{t}||+1)$                 \\
        Fall down penalty & -1                                             \\
        \hline
        \multicolumn{2}{c}{Reward design 5}                                \\
        \hline
        Pose reward       & $\exp(-||\boldsymbol{p}_{t}||)$                \\
        Fall down penalty & -1                                             \\
        \hline
    \end{tabular}
\end{table}
In the experiment, we trained the policy to make the robot transition from an arbitrary initial pose to an upright pose
(See Fig.~\ref{fig1-1}) and tested its performance using 10 different initial poses.
We built a simulation environment using MuJoCo~\cite{IROS2012_Todorov} and
trained the policy for 1 million iterations.
To facilitate faster convergence, we performed curriculum training,
starting from an initial pose near the upright position and
gradually increasing the distance from the upright pose.
We did not perform parallel simulations to train the policy and used
only one simulated robot to collect data during training.
In all experiments, the trained controller operates at \SI{10}{Hz}.
See the appendix for the hyperparameters used in the experiment.

In the simulation, we evaluated the proposed method comparing it with:
\begin{itemize}
    \item A policy learned using manually designed reward functions:
          Table~\ref{tab6-1} shows the five different reward functions designed based on the task's learnability.
    \item A policy learned using conventional methods:
          We compared the proposed QRSAC-Lagrangian with conventional algorithms such as SAC-Lagrangian~\cite{CoRL2020_Ha},
          PPO-Lagrangian~\cite{Arxiv2019_Achiam}, and CaT~\cite{ICRA2024_Elliot}.
\end{itemize}
Additionally, we assessed the robustness of the policy by testing it on terrains different from the trained environment.
Furthermore, we conducted an ablation study to evaluate the efficacy of each constraint designed in the previous section.

For the real robot experiment, we ran the policy trained with the proposed method without any fine-tuning on real data
and evaluated its performance.

\subsection{Evaluation Results}
\begin{figure*}[tb]
    \centering
    \begin{tabular}{c:cccccc}
        \includegraphics[width=0.245\columnwidth]{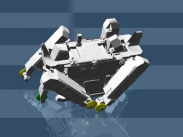}        &
        \includegraphics[width=0.245\columnwidth]{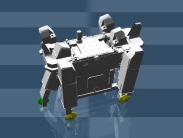} &
        \includegraphics[width=0.245\columnwidth]{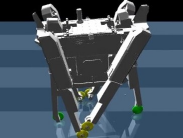} &
        \includegraphics[width=0.245\columnwidth]{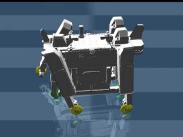} &
        \includegraphics[width=0.245\columnwidth]{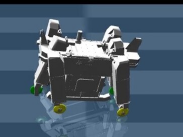} &
        \includegraphics[width=0.245\columnwidth]{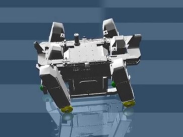} &
        \includegraphics[width=0.245\columnwidth]{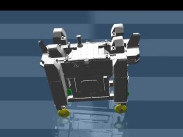}                                                                                                                                                           \\
                                                                                    & (0.679)                                                                             & (0.477)  & (0.750)  & (0.670)  & (0.640)  & (0.930)    \\
        \includegraphics[width=0.245\columnwidth]{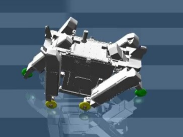}        &
        \includegraphics[width=0.245\columnwidth]{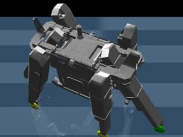} &
        \includegraphics[width=0.245\columnwidth]{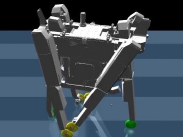} &
        \includegraphics[width=0.245\columnwidth]{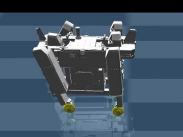} &
        \includegraphics[width=0.245\columnwidth]{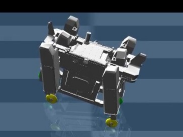} &
        \includegraphics[width=0.245\columnwidth]{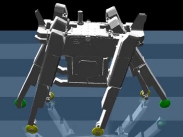} &
        \includegraphics[width=0.245\columnwidth]{figures/car_pose1.pdf}                                                                                                                                                           \\
                                                                                    & (0.420)                                                                             & (0.476)  & (0.751)  & (0.671)  & (0.458)  & (0.930)    \\
                                                                                    & Design 1                                                                            & Design 2 & Design 3 & Design 4 & Design 5 & CaR (ours) \\
        Initial pose                                                                & \multicolumn{6}{c}{Final pose (The number in parenthesis is the final pose score.)}                                                          \\
    \end{tabular}
    \caption{Final poses of the robot. Leftmost figure shows the initial pose of each row. The policy trained with manually designed rewards fails to transition to the upright pose.
        In contrast, our proposed method (CaR) succeeds in standing up.}
    \label{fig6-2}
\end{figure*}
Fig.~\ref{fig6-2} shows
the final pose of the robot when running policies trained with
manually designed rewards and the proposed method\footnote{In this experiment, the policy with the proposed method is trained only with the final pose contraint and the fall-down constraint for a fair comparison.}.
The pose score $1/(||\boldsymbol{p}_{t}||+1)$ is also shown in parenthesis.
From the figure, we can confirm that CaR succeeds in transitioning to the target pose.
In contrast, policies trained with manually designed rewards fail to transition to the target pose.
Among the policies trained with manually designed rewards,
Design 3 exhibits the highest final pose score; however, this score remains lower compared to the proposed method.
We do not claim that there is no reward function that can achieve this task.
However, from the experiment, we can confirm that designing a reward function is not a straightforward task,
and the proposed method is effective even in such situations.

\begin{figure}[tb]
    \centering
    \begin{minipage}{0.49\columnwidth}
        \includegraphics[width=\columnwidth]{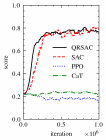}
        \subcaption{Algorithm learning curve.}
        \label{fig6-3a}
    \end{minipage}
    \begin{minipage}{0.482\columnwidth}
        \includegraphics[width=\columnwidth]{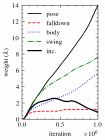}
        \subcaption{Weight parameters.}
        \label{fig6-3b}
    \end{minipage}
    \caption{Algorithm learning curve and tuning history of weight parameters.
        The learning curve shows the average of 5 runs initialized with different random seeds.
        The weight of the pose parameter shown on the right is the weight of the left-front hip joint
        (Other weight parameters for pose constraints also exhibited similar changes).}
\end{figure}
Fig.~\ref{fig6-3a} shows the learning curve of the proposed algorithm and comparison algorithms in the task.
From the figure, we can confirm that the proposed QRSAC-Lagrangian achieves faster and more robust convergence
compared to conventional algorithms. PPO-Lagrangian and CaT failed to learn the task.
SAC-Lagrangian achieves similar final performance, but its convergence is slower compared to QRSAC-Lagrangian.
The difference between SAC-Lagrangian and QRSAC-Lagrangian lies solely in the underlying reinforcement learning algorithm.
This result suggest that the quantile function introduced in QRSAC is effective in the setting of CaR,
where the distrubution of the target Q value dynamically changes during training.
Also, refer to the appendix for results confirming similar findings in the classic inverted pendulum task.
Fig.~\ref{fig6-3b} shows the tuning results of weight parameters conducted by the proposed algorithm.
We can confirm that weight parameters were tuned dynamically during training.
The weight for fall-down constraint converged quickly because the robot learned to avoid falling down at the
beginning of training. In contrast, the weights for the pose constraint, body constraint, and swing constraint gradually incresed
during the course of training. This result indicates that these constraints were challenging for the policy to meet.
The weight for inclination constraint increased from the beginning to the middle of training but
decreased towards the end. This is because the policy succeeded in keeping the final pose
parallel to the floor by the middle of training,
and the algorithm decreased its weight towards the end to prioritize the pose, body, and swing constraints more.

\begin{figure}[tb]
    \centering
    \begin{tabular}{ccc}
        \includegraphics[width=0.28\columnwidth]{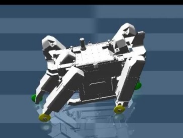} & \includegraphics[width=0.28\columnwidth]{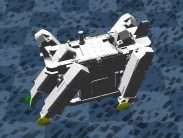} & \includegraphics[width=0.28\columnwidth]{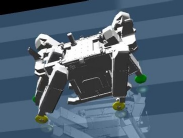} \\
        \includegraphics[width=0.28\columnwidth]{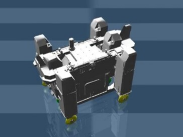}   & \includegraphics[width=0.28\columnwidth]{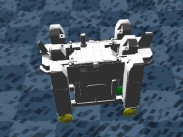}   & \includegraphics[width=0.28\columnwidth]{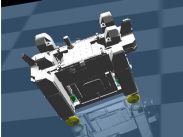}   \\
        (0.902)                                                            & (0.902)                                                             & (0.903)                                                             \\
        Flat                                                               & Rough                                                               & Slope                                                               \\
    \end{tabular}
    \caption{Policy execution results on different types of terrain.
        (Top) Initial pose. (Bottom) Final pose.
        Policy is only trained in a flat terrain environment but works on different types of terrain.
        The score in parenthesis is the final pose score.}
    \label{fig6-4}
\end{figure}
Fig.~\ref{fig6-4} shows the execution results of the trained policy in three different environments.
The flat terrain environment is the trained environment.
The rough terrain environment has a random height of up to \SI{0.05}{m}.
The slope terrain environment has a slope of \SI{10}{deg}.
From the figure, we can confirm the robustness of the policy.
The trained policy succeeds in completing the task even in
environments different from the original trained environment.

\begin{table}[t]
    \caption{Ablation results. Each value is the average of 10 runs.
        Contact is the ratio of robot's body making contact with the floor during the task.
        Max $\omega_{\text{joint}}$ is the maximum angular velocity of the leg joint recorded during the task.
        $\theta_{T,\text{roll}}$ and $\theta_{T,\text{pitch}}$ are the roll and pitch angles of the robot at the final timestep.
        $\uparrow$ and $\downarrow$ indicate that a higher value is better and a lower value is better, respectively.
        (P: Pose constraint, F: Fall-down constraint, C: Contact constraint, S: Swing constraint, I: Inclination constraint)}
    \label{tab6-2}
    \centering
    \begin{tabular}{l|c|c|c|c|c}
                  & Pose             & Contact           & Max $\omega_{\text{joint}}$ & $\theta_{T}^{\text{roll}}$ & $\theta_{T}^{\text{pitch}}$ \\
                  & score $\uparrow$ & [\%] $\downarrow$ & [rad/s] $\downarrow$        & [deg] $\downarrow$         & [deg] $\downarrow$          \\
        P+F       & \textbf{0.925}   & 0.5               & 3.038                       & 3.896                      & 5.672                       \\
        P+F+C     & 0.920            & \textbf{0.1}      & 8.937                       & 6.933                      & 3.782                       \\
        P+F+C+S   & 0.901            & 1.0               & \textbf{1.136}              & 3.839                      & 4.698                       \\
        P+F+C+S+I & 0.896            & 0.6               & 1.673                       & \textbf{0.057}             & \bf{0.057}
    \end{tabular}
\end{table}
Table~\ref{tab6-2} shows the results of the ablation study on the designed constraint functions.
From the table, we can confirm that each constraint succeeded in improving the performance of the corresponding metric.
Additionally, while it is challenging to perfectly satisfy each constraint,
the average performance of the policy trained with all five constraints appears reasonable.
Based on these results, we used the policy trained with all five constraints in the real robot experiment.

\begin{figure}[tb]
    \centering
    \includegraphics[width=1.0\columnwidth]{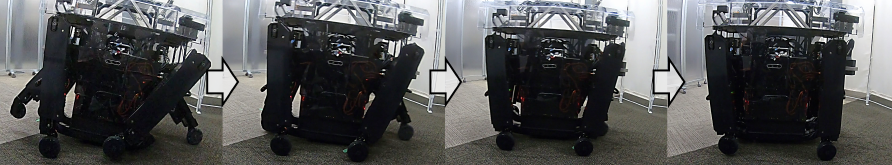} \\
    \includegraphics[width=1.0\columnwidth]{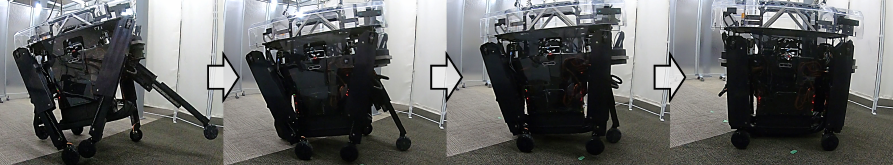}
    \caption{Policy execution result on the real robot. (Top) Pose used in simulation. (Bottom) Challenging pose with expanded knees.}
    \label{fig6-5}
\end{figure}
Fig.~\ref{fig6-5} shows the execution result of the trained policy in a real environment.
From the figure, we can confirm that the policy succeeds in completing the task on the real robot.
We tested 10 poses used in the simulation and 9 challenging poses with expanded knees,
and confirmed that the policy successfully transitions to the upright pose from all poses.

\expandafter\ifx\csname ifdraft\endcsname\relax
\bibliographystyle{IEEEtran}
\bibliography{ref_iros2025}
\end{document}
\fi
\expandafter\ifx\csname ifdraft\endcsname\relax
    \documentclass[letterpaper, 10 pt, conference]{ieeeconf}
    \overrideIEEEmargins

\begin{document}
\fi

\section{CONCLUSION AND FUTURE WORK}
In this paper, we introduced the concept of Constraints as Rewards (CaR) to mitigate the extensive 
trial-and-error involved in tuning a reward function.
Instead of designing a reward function, CaR composes the task objective solely from constraint functions.
In this approach, unlike conventional reward function design, we do not need to manually tune
the weights among task objectives; the weights are adjusted automatically during the training process.
In addition, to simplify the design of constraint functions,
we proposed four specific designs that provide an intuitive interpretation of the task objective.
Furthermore, we introduced QRSAC-Lagrangian algorithm to solve this reinforcement learning problem
with constraints.
We demonstrated the effectiveness of our method by applying it to the standing-up motion generation task of
a six-wheeled-telescopic-legged  
\expandafter\ifx\csname ifanonymous\endcsname\relax
robot.
\else
robot, Tachyon 3.
\fi
While this task is challenging to learn with manually designed reward functions,
our proposed method enabled the robot to effectively learn the target behavior.

We believe that CaR is effective for a wide range of robotic tasks. 
However, CaR requires the task to be expressed 
solely in terms of constraint functions. 
Therefore, it might be challenging to apply in tasks where
pure maximization is required (e.g., make the robot walk as fast as possible).
In such cases, it would be beneficial to combine reward functions with constraint functions.
Exploring an effective objective design method for such tasks could be a promising direction for future work.

\expandafter\ifx\csname ifdraft\endcsname\relax
\end{document}
\fi
\expandafter\ifx\csname ifanonymous\endcsname\relax
\else
\expandafter\ifx\csname ifdraft\endcsname\relax
    \documentclass[letterpaper, 10 pt, conference]{ieeeconf}
    \overrideIEEEmargins
    \usepackage{siunitx}
    \usepackage[dvipdfmx]{graphicx}
    \usepackage{subcaption}
\begin{document}
\fi

\section*{Acknowledgment}
We would like to thank Takuma Seno, Sotaro Katayama and Michael Yeung for their helpful comments and feedbacks
during the preparation of this manuscript.

\expandafter\ifx\csname ifdraft\endcsname\relax
\end{document}
\fi
\fi

\addtolength{\textheight}{0cm}
\bibliographystyle{IEEEtran}
\bibliography{ref_iros2025}

\expandafter\ifx\csname ifdraft\endcsname\relax
    \documentclass[letterpaper, 10 pt, conference]{ieeeconf}
    \overrideIEEEmargins
    \usepackage{siunitx}
    \usepackage[dvipdfmx]{graphicx}
    \usepackage{subcaption}
    \usepackage{amssymb}
    \usepackage{amsmath}
    \usepackage{amsfonts}
    \usepackage{multirow}
\begin{document}
\fi

\appendices
\section{Constraint function derivation}
In the derivation, we assume discrete state and action sets but
the results can also be applied to continuous settings.
\subsection{Derivation of timestep probability constraint}
By substituting eq. (\ref{eq3-1}) into the optimization objective, we get:
\begin{align*}
      & \mathbb{E}_{\pi}[\sum_{t=0}^{T}\gamma^{t}g(s_t, a_t)]                                                                             \\
    = & \mathbb{E}_{\pi}[\gamma^{t'}(p_{\epsilon}-\mathbf{1}_{s\in\mathcal{S}',a\in\mathcal{A}'})]\quad(\because\ g(s_t,a_t)=0\ t\neq t') \\
    = & \gamma^{t'}(p_{\epsilon}-\mathbb{E}_{\pi}[\mathbf{1}_{s\in\mathcal{S}',a\in\mathcal{A}'}])                                        \\
    = & \gamma^{t'}(p_{\epsilon}-P_{\pi}(s\in\mathcal{S}',a\in\mathcal{A}')).
\end{align*}
Therefore, the constraint can be expressed as follows:
\begin{equation*}
    \gamma^{t'}(p_{\epsilon}-P_{\pi}(s\in\mathcal{S}',a\in\mathcal{A}')) \geq 0 \rightarrow
    p_{\epsilon} \geq P_{\pi}(s\in\mathcal{S}',a\in\mathcal{A}').
\end{equation*}

\subsection{Derivation of timestep value constraint}
Similar to the above derivation,
by substituting eq.\ref{eq3-2} into the optimization objective, we get:
\begin{equation*}
    \gamma^{t'}(\epsilon-\mathbb{E}_{\pi}[\hat{g}(s_{t'},a_{t'})]) \geq 0 \rightarrow
    \epsilon \geq \mathbb{E}_{\pi}[\hat{g}(s_{t'},a_{t'})].
\end{equation*}

\subsection{Derivation of episode probability constraint}
First, we reformulate the optimization objective as follows:
\begin{align*}
      & \mathbb{E}_{\pi}[\sum_{t=0}^{T}\gamma^{t}g(s_t, a_t)]                                                                                                   \\
    = & \sum_{t=0}^{T}\sum_{s_0\in\mathcal{S},\dots,a_T\in\mathcal{A}}p(s_0,a_0,\dots,s_T,a_T)\gamma^{t}g(s_t,a_t)                                              \\
    = & \sum_{t=0}^{T}\sum_{s\in\mathcal{S}}\sum_{a\in\mathcal{A}}p_{t}^{\pi}(s,a)\gamma^{t}g(s,a)                                                              \\
    = & \sum_{s\in\mathcal{S}}\sum_{a\in\mathcal{A}}g(s,a)\sum_{t=0}^{T}p^{\pi}_t(s,a)\gamma^{t}                                                                \\
    = & \frac{1-\gamma^{T+1}}{1-\gamma}\sum_{s\in\mathcal{S}}\sum_{a\in\mathcal{A}}g(s,a)\sum_{t=0}^{T}p^{\pi}_t(s,a)\frac{1-\gamma}{1-\gamma^{T+1}}\gamma^{t}.
\end{align*}
Here, $p^{\pi}_t(s,a)$ is the state-action probability distribution at time $t$ under policy $\pi$.
By defining the discounted state-action distribution as:
\begin{equation*}
    p_{\pi,\gamma}(s,a)\triangleq\sum_{t=0}^{T}p^{\pi}_t(s,a)\frac{1-\gamma}{1-\gamma^{T+1}}\gamma^{t},
\end{equation*}
the original expectation can be rewritten as:
\begin{equation*}
    \mathbb{E}_{\pi}\biggr[\sum_{t=0}^{T}\gamma^{t}g(s_t, a_t)\biggl]\propto\mathbb{E}_{(s,a)\sim p_{\pi,\gamma}}\biggr[g(s, a)\biggl]
\end{equation*}
Next, by substituting eq. (\ref{eq3-3}), we get:
\begin{align*}
    \mathbb{E}_{\pi}[\sum_{t=0}^{T}\gamma^{t}g(s_t, a_t)] & \propto\mathbb{E}_{(s,a)\sim p_{\pi,\gamma}}\biggr[g(s, a)\biggl]                                                  \\
                                                          & = \mathbb{E}_{(s,a)\sim p_{\pi,\gamma}}\biggr[(p_{\epsilon}-\mathbf{1}_{s\in\mathcal{S}',a\in\mathcal{A}'})\biggl] \\
                                                          & = (p_{\epsilon}-\mathbb{E}_{(s,a)\sim p_{\pi,\gamma}}\biggr[\mathbf{1}_{s\in\mathcal{S}',a\in\mathcal{A}'}\biggl]) \\
                                                          & = p_{\epsilon}-P_{\pi,\gamma}(s\in\mathcal{S}',a\in\mathcal{A}').
\end{align*}
Therefore, the constraint can be expressed as follows:
\begin{equation*}
    p_{\epsilon}-P_{\pi,\gamma}(s\in\mathcal{S}',a\in\mathcal{A}') \geq 0. \rightarrow
    p_{\epsilon} \geq P_{\pi,\gamma}(s\in\mathcal{S}',a\in\mathcal{A}').
\end{equation*}

\subsection{Derivation of episode value constraint}
Similar to the above derivation,
by sustituting eq. (\ref{eq3-4}) into the optimization objective, we get:
\begin{equation*}
    \frac{1-\gamma^{T+1}}{1-\gamma}(\epsilon-\mathbb{E}_{\pi,\gamma}[\hat{g}(s,a)]) \geq 0 \rightarrow
    \epsilon \geq \mathbb{E}_{\pi,\gamma}[\hat{g}(s,a)].
\end{equation*}
Here, $\mathbb{E}_{\pi,\gamma}$ is the expectation under the discounted state-action distribution $p_{\pi,\gamma}$.

\section{Hyperparameters and experimental settings}
\begin{table}[t]
    \caption{QRSAC-Lagrangian hyperparameters}
    \label{tabA-1}
    \centering
    \begin{tabular}{c|c}
        Adam parameters                               & $\beta_{1}=0.9$, $\beta_{2}=0.999$ \\
        Model learning rate                           & $3.0\times10^{^-4}$                \\
        Multiplier learning rate ($\alpha_{\lambda}$) & 0.1                                \\
        Multiplier update interval ($d$)              & 5000                               \\
        Discount factor $\gamma$                      & 0.99                               \\
        Batch size                                    & 256                                \\
        Quantile points                               & 32                                 \\
        Network update coefficient $\tau$             & 0.005                              \\
    \end{tabular}
\end{table}

\begin{table}[t]
    \caption{Network structure of quantile function}
    \label{tabA-2}
    \centering
    \begin{tabular}{|c|c|}
        \hline
        Input layer    & state and action input                     \\
        \hline
        Middle layer 1 & Fully-connected followed by relu (256 dim) \\
        \hline
        Middle layer 2 & Fully-connected followed by relu (256 dim) \\
        \hline
        Middle layer 3 & Fully-connected followed by relu (256 dim) \\
        \hline
        Output layer   & Quantile values (32 dim)                   \\
        \hline
    \end{tabular}
\end{table}

\begin{table}[t]
    \caption{Noise applied to model input during training.
        $\mathcal{N}(\mu,\sigma^2)$ is a Gaussian distribution with mean $\mu$ and variance $\sigma^2$.}
    \label{tabA-3}
    \centering
    \begin{tabular}{c|c}
        Joint angle noise              & $\mathcal{N}(\mu=0, \sigma^2=0.005)$ \\
        Knee position noise            & $\mathcal{N}(\mu=0, \sigma^2=0.005)$ \\
        Joint angular velocity noise   & $\mathcal{N}(\mu=0, \sigma^2=0.01)$  \\
        Knee velocity noise            & $\mathcal{N}(\mu=0, \sigma^2=0.01)$  \\
        Robot's angular velocity noise & $\mathcal{N}(\mu=0, \sigma^2=0.01)$  \\
        Robot's accelaration noise     & $\mathcal{N}(\mu=0, \sigma^2=0.1)$   \\
        Timestep noise                 & $\mathcal{N}(\mu=0, \sigma^2=0.1)$   \\
    \end{tabular}
\end{table}
Hyperparameters and settings used in the experiments are listed in Table~\ref{tabA-1},Table~\ref{tabA-2}, and Table~\ref{tabA-3}.

\section{Extra experimental results}
We present the evaluation results of the proposed algorithm in
the classic inverted pendulum task implemented in Gym~\cite{Arxiv2016_Brockman}.
We trained the model using two different constraint functions:
\begin{equation}
    \label{eqA-1}
    g(s,a) = 10^{-2}-|\theta_{p}|
\end{equation}
and
\begin{equation}
    \label{eqA-2}
    g(s,a) =
    \begin{cases}
        0 \quad (t \neq T) \\
        10^{-2}-|\theta_{p}| \quad (t=T).
    \end{cases}
\end{equation}
Here, $\theta_{p}$ is the angle of the pendulum.
\begin{figure}[t]
    \centering
    \includegraphics[width=\columnwidth]{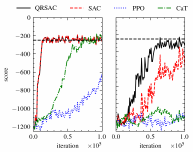}
    \caption{Learning curve of each algorithm trained with eq.~(\ref{eqA-1}) (Left) and eq.~(\ref{eqA-2}) (Right). Dotted line is the score when trained with original reward function.}
    \label{figA-1}
\end{figure}
Fig.~\ref{figA-1} shows the learning curves during training.
From the figure, we can confirm that conventional PPO-Lagrangian and CaT failed to learn with the constraint function in eq.~(\ref{eqA-2}).
This result suggests that PPO-Lagrangian and CaT are not effective in environments with constraint functions of the form in eq.~(\ref{eqA-2}).
Conversely, SAC-Lagrangian and QRSAC-Lagrangian succeeded in learning in both settings.
As in the experiment section, QRSAC-Lagrangian converged faster than SAC-Lagrangian.

\expandafter\ifx\csname ifdraft\endcsname\relax
\bibliographystyle{IEEEtran}
\bibliography{ref_iros2025}
\end{document}
\fi

\end{document}